\documentclass[conference]{IEEEtran}
\IEEEoverridecommandlockouts
\usepackage{cite}
\usepackage{amsmath,amssymb,amsfonts}
\usepackage{graphicx}
\usepackage{textcomp}
\usepackage{xcolor}
\usepackage{algorithm}

\usepackage{algpseudocode}
\usepackage{booktabs} 
\usepackage{threeparttable} 
\usepackage[utf8]{inputenc}
\def\BibTeX{{\rm B\kern-.05em{\sc i\kern-.025em b}\kern-.08em
    T\kern-.1667em\lower.7ex\hbox{E}\kern-.125emX}}
\begin{document}

\title{Differential Private Federated Transfer Learning for Mental Health Monitoring in Everyday Settings: A Case Study on Stress Detection
\thanks{*Authors equally contributed.}
 }

\author{\IEEEauthorblockN{$^1$Ziyu Wang$^*$, $^1$Zhongqi Yang$^*$, $^{1,3}$Iman Azimi, and $^{1,2,3}$Amir M. Rahmani\\
\textit{$^1$Department of Computer Science, University of California, Irvine}\\
\textit{$^2$School of Nursing, University of California, Irvine}\\
\textit{$^3$Institute for Future Health, University of California, Irvine}\\
}
\IEEEauthorblockA{
\{ziyuw31, zhongqy4, azimii, a.rahmani\}@uci.edu}}

\maketitle

\begin{abstract}
Mental health conditions, prevalent across various demographics, necessitate efficient monitoring to mitigate their adverse impacts on life quality. 
The surge in data-driven methodologies for mental health monitoring has underscored the importance of privacy-preserving techniques in handling sensitive health data.
Despite strides in federated learning for mental health monitoring, existing approaches struggle with vulnerabilities to certain cyber-attacks and data insufficiency in real-world applications. 
In this paper, we introduce a differential private federated transfer learning framework for mental health monitoring to enhance data privacy and enrich data sufficiency.
To accomplish this, we integrate federated learning with two pivotal elements: (1) differential privacy, achieved by introducing noise into the updates, and (2) transfer learning, employing a pre-trained universal model to adeptly address issues of data imbalance and insufficiency.
 We evaluate the framework by a case study on stress detection, employing a dataset of physiological and contextual data from a longitudinal study.
Our finding show that the proposed approach can attain a 10\% boost in accuracy and a 21\% enhancement in recall, while ensuring privacy protection.
\end{abstract}


\section{Introduction}

Mental health concerns impact a diverse demographic cross-section of the population~\cite{prince2007no}. 
The prolonged effects of these mental health conditions can have detrimental impacts on physical health~\cite{cheng2024shortterm, yang2024chatdiet}, psychological states, and overall quality of life~\cite{zhou2024glumarker, cheng2023saic, zhou2024crossgp}. In response, the field of mental health monitoring has increasingly turned to data-driven methodologies~\cite{bucci2019digital, alikhani2024seal}. 
These methods, utilizing advanced analytics and machine learning algorithms, provide precise, real-time assessments of various mental health states~\cite{garcia2018mental}.
However, the ascendancy of data-driven techniques in mental health monitoring brings forth pressing concerns regarding data privacy \cite{Kanduri2024}.
The sensitive nature of the data involved demands methodologies that are not just accurate and efficient in monitoring mental health but also stringently protective of individual privacy \cite{wang2020guardhealth, yang2022zebra, alikhani2023dynafuse}. 


Recent studies in mental health monitoring (e.g., mood, loneliness, depression, stress) increasingly adopt 
federated learning to enhance privacy \cite{xu2021fedmood,alahmadi2023privacy, chen2020fedhealth,suruliraj2022federated,qirtas2022privacy, can2021privacy, yao2020privacy}, showing promise in using local data for model training. For example, \cite{suruliraj2022federated} developed a framework for depression detection using smartphone data (location, accelerometer, call logs) for training on devices, although vulnerable to privacy breaches like membership inference attacks. \cite{qirtas2022privacy} and \cite{can2021privacy, sharifi2023phenotyping} similarly leveraged federated learning for loneliness and stress detection, utilizing sensor and PPG signal data from smartphones and wearables, respectively, to update models locally, thus preserving personal data privacy.

Although existing approaches have shown success in enhancing privacy in certain domains, they still exhibit vulnerabilities to a range of cyber threats, notably backdoor and inference attacks. 
The absence of more sophisticated privacy-preserving mechanisms, such as those offered by differential privacy, represents a gap in the current approaches for mental health monitoring. 
Furthermore, a significant hurdle in the practical deployment of federated learning in mental health studies stems from the reliance on limited data sources \cite{hu2020dasgil, cheng2024efflex}. When models are trained locally in such studies, there's often a marked inconsistency in the diversity and volume of data, as noted by Dixit et al. \cite{dixit2020managing}. This variability can lead to inadequate training of decentralized models, especially when real-world data lacks the comprehensiveness or balance needed for effective training \cite{cheng2024vetrass}. Therefore, resolving this issue is imperative for the effective application of federated learning in practical scenarios.

In this study, we propose a differential private federated transfer learning framework for mental health monitoring.
We incorporate a differential privacy mechanism into federated learning by introducing noise into the combined updates prior to their transmission back to the central server.
Furthermore, we utilize transfer learning to tackle data insufficiency and imbalance in stress detection as a case study. Specifically, we commence by pre-training a universal model on an extensive, non-sensitive dataset, subsequently refining it with individual user data on-device.
We evaluate the proposed framework via a case study on stress detection on a dataset from a longitudinal study. 
The dataset includes physiological and contextual data gathered from Samsung Galaxy Active 2 Watches and smartphones from 54 individuals. The label stress levels of these individuals were collected through multiple daily ecological momentary assessments (EMAs) conducted via their smartphones.



\section{Methodology}


\begin{figure}[t]
\centering
\includegraphics[width=0.38\textwidth]{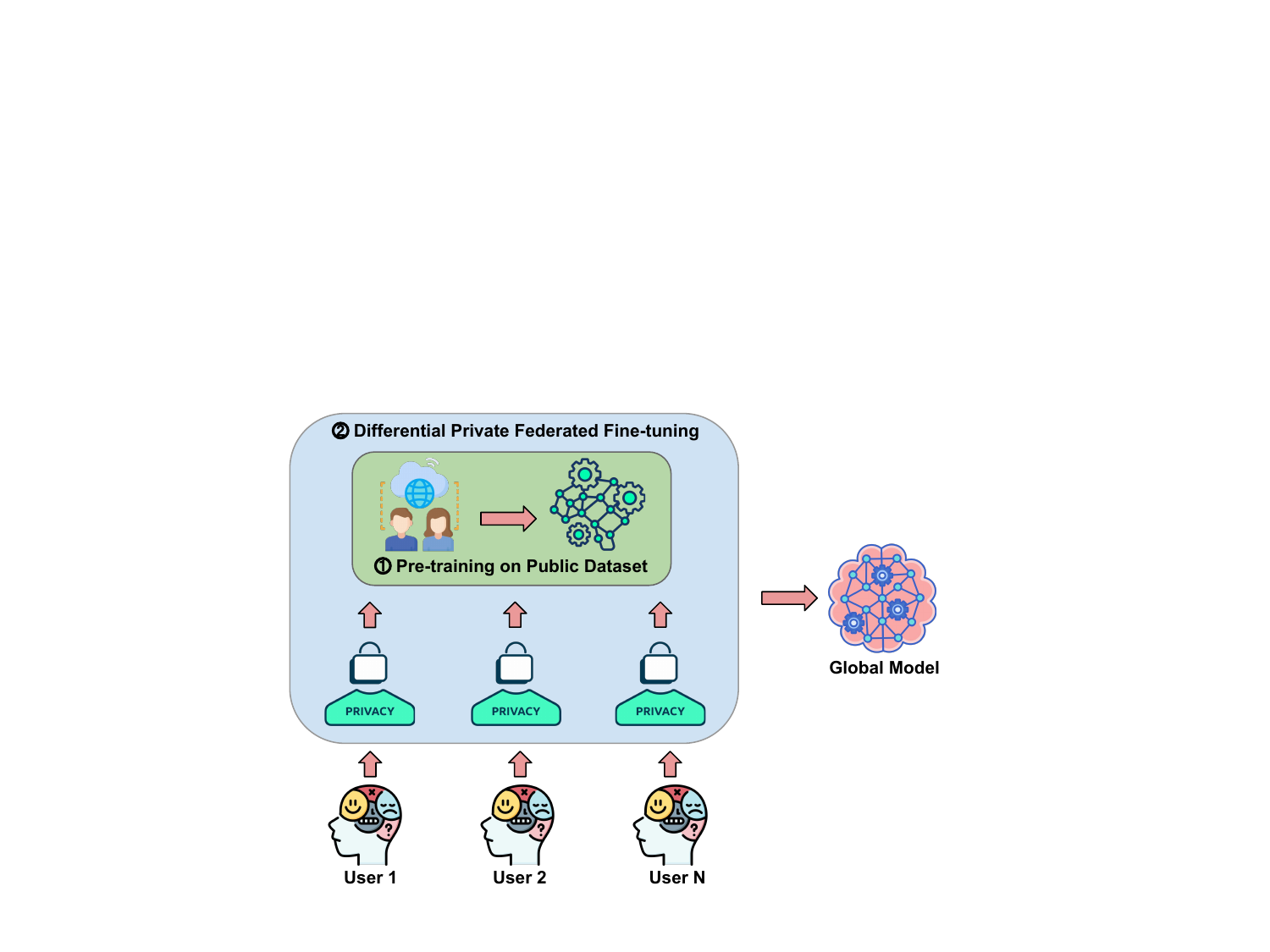}
\vspace{-6pt}
\caption{Overview of the proposed framework.}
\label{fig:overview}
\vspace{-15pt}
\end{figure}

Our methodology addresses data scarcity and privacy challenges in mental health monitoring by integrating transfer learning with federated learning, enhanced by differential privacy techniques.
We initially pre-train a foundational model on a comprehensive public dataset to capture broad health trends. This model is then fine-tuned on sparse, user-specific data in a distributed manner, improving accuracy with limited data. Differential privacy is applied during weight updates to protect individual data privacy. This combined approach effectively balances privacy concerns with the need for precise health monitoring. Our methodology is depicted in Fig. \ref{fig:overview}.

\subsection{Differential Private Federated Transfer Learning Framework for Mental Health Monitoring}

The learning process is detailed in Algorithm \ref{alg:framework}, and we provide an in-depth explanation of each module in the subsequent sections.

\begin{algorithm}
\caption{Differential Private Federated Transfer Learning for Mental Health Monitoring}\label{alg:framework}.
\begin{algorithmic}[1]
\State \textbf{Pre-training Phase:}
\State $\mathcal{D}_{pre-train} \gets$ Load publicly available dataset
\State Initialize model $\mathcal{M}$
\For{each batch $(x_i, y_i)$ in $\mathcal{D}_{pre-train}$}
    \State Update $\mathcal{M}$ to minimize $\mathcal{L}_{pre-train}(\mathcal{M})$
\EndFor

\State \textbf{Federated Fine-Tuning Phase:}
\For{each client $k$ in the federated network}
    \State Distribute pre-trained model $\mathcal{M}$ to client $k$
    \State $\mathcal{D}_{\text{user}, k} \gets$ Load user-specific dataset for client $k$
    \For{each batch $(x'_j, y'_j)$ in $\mathcal{D}_{\text{user}, k}$}
        \State Fine-tune $\mathcal{M}$ on $(x'_j, y'_j)$
    \EndFor
    \State Compute gradients and apply gradient clipping
    \State Add differential privacy noise $\text{Lap}\left(\frac{\Delta f}{\epsilon}\right)$ to gradients
    \State Send updated model $\mathcal{M}_k$ to server
\EndFor

\State \textbf{Global Model Aggregation:}
\State Initialize global model $\mathcal{M}_{\text{global}}$
\State $\mathcal{M}_{\text{global}} \gets$ Aggregate updates $\mathcal{M}_k$ from all clients

\State \textbf{return} $\mathcal{M}_{\text{global}}$

\end{algorithmic}

\end{algorithm}


\subsubsection{Pre-training}

In real-world scenarios, gathering extensive, high-quality data for mental health studies is often fraught with difficulties. Participants in such studies may not consistently engage, leading to gaps or irregularities in the data. Additionally, the collected data often contains artifacts and noise, further complicating its utility and quality. These factors combine to create a scenario where the available data is scarce, inconsistent, and noisy.

We initiate our approach by pre-training our model $\mathcal{M}$ on a comprehensive publicly available dataset $\mathcal{D}{\text{pt}}$. This dataset, rich in both quantity and quality, comprises $N$ pairs of well-curated features and labels:
\begin{equation}
{D_{pre-train}} = {(x_1, y_1), (x_2, y_2), ..., (x_N, y_N)}.
\end{equation}

The pre-training phase utilizes the binary cross-entropy loss, and is aimed at learning generalized mental health patterns:

\vspace{-12pt}
\begin{equation}
\label{eqn:transloss}
    \mathcal{L} = -\frac{1}{N} \sum_{i=1}^{N} \left[ y_i \cdot \log(\mathcal{M}(x_i)) + (1 - y_i) \cdot \log(1 - \mathcal{M}(x_i)) \right].
\end{equation}



\subsubsection{Differential Private Federated Fine-tuning}

To tackle the critical privacy concerns in mental health monitoring, our approach incorporates differential privacy (DP) within the federated learning paradigm, specifically through the Federated Averaging (FedAvg) algorithm \cite{mcmahan2017communication} enhanced with Laplacian noise.
This implementation is key to defending against sophisticated privacy attacks against traditional federated learning frameworks, including model inversion \cite{fredrikson2015model}, membership inference \cite{nasr2018comprehensive}, and backdoor \cite{bagdasaryan2020backdoor}, which pose risks to user data confidentiality.


\paragraph{Federated Averaging} FedAvg is a widely-used algorithm in federated learning that involves averaging model updates (weights) from multiple clients. In our implementation, each client represents a participant in the mental health monitoring study, first trains a local model using their private data. The local models are then aggregated to update the global model. This process is formalized as:

\vspace{-5pt}
\begin{equation}
\vspace{-5pt}
\mathcal{M}_{\text{global}} = \frac{1}{K} \sum_{k=1}^{K} \mathcal{M}_{k},
\end{equation}

where $\mathcal{M}_{\text{global}}$ represents the global model, $\mathcal{M}_{k}$ is the model trained on the $k$-th client, and $K$ is the total number of clients.

\paragraph{Laplacian Noise for Differential Privacy} To ensure differential privacy, we add Laplacian noise to the model updates during aggregation. The noise is calibrated to the sensitivity of the model's gradients and the privacy budget $\epsilon$. For a model parameter $\theta$, the update with Laplacian noise is:
\vspace{-14pt}

\begin{equation}
\theta_{\text{updated}} = \theta + \text{Lap}\left(\frac{\Delta f}{\epsilon}\right),
\end{equation}
where $\Delta f$ denotes the sensitivity of the function (gradient norm), and $\text{Lap}(\cdot)$ represents the Laplacian noise.

\paragraph{Fine-tuning} Following the pre-training phase, the model $\mathcal{M}$ utilize private, user-specific datasets, represented as $D_{fine-tune} = \{(x'_j, y'_j)\}_{j=1}^{M}$ to fine-tune. 
This process occurs within a federated learning framework and is essential for adapting the generalized model to the specific mental health profiles and solve the data scarcity issue of the private dataset. 
During fine-tuning, the model parameters are adjusted according to each user's data. The fine-tuning employs the same loss function, as Eq. \ref{eqn:transloss}, used in the pre-training phase, ensuring consistency in the optimization approach across both stages of model development.

\subsection{Case Study on Stress Detection}
The efficacy of our framework is showcased through a case study in stress detection, leveraging data from an extensive longitudinal study \cite{tazarv2023}. This study engaged distinct groups of participants across its two phases, with 30 individuals in the initial phase and 24 in the subsequent phase, including both undergraduate and graduate students. Data collection occurred over two periods: from June 2020 to June 2021 and from March 2022 to May 2023, resulting in 109,586 and 23,012 refined samples for each period, respectively.
The dataset from the first phase, $D_{pre-train}$, was critical for the pre-training process, and the dataset from the second phase, $D_{fine-tune}$, was essential for fine-tuning. 
Conducted using our ZotCare health monitoring system \cite{zotcare}, this strategy leveraged the comprehensive dataset from the first phase for initial model training, while the more constrained dataset from the second phase was used for precise model enhancement, effectively tackling the challenges posed by data scarcity.

\subsubsection{Dataset}
The collected dataset comprises raw PPG and motion data, along with partial annotations for stress levels, emotions, and physical activity through EMA at semi-random intervals. To comprehensively process the PPG signals, we employ the HeartPy library \cite{van2019heartpy}, extracting 12 specific features from both electrical activations and pressure waveforms in the dataset.
These features\footnote{BPM, IBI, SDNN, SDSD, RMSSD, pNN20, pNN50, MAD, SD1, SD2, S, SD1/SD2, and BR1.} 
are heart rate (HR) and heart rate variability (HRV) measures.
Stress levels equal to or less than 2 are categorized as 'unstressed,' while higher values indicate 'stressed' individuals.

\subsubsection{Stress Detection Model}


To develop our stress detection model, we began with a pre-processing stage to refine bio-signals from wearable devices. Initial data cleaning eliminated erroneous readings by using motion data to filter out noise and artifacts, utilizing a bandpass Butterworth filter specific to PPG signal frequencies. This was followed by a moving average technique to smooth the data, minimizing motion-related distortions. We then applied min-max normalization, scaling feature values uniformly between 0 and 1, to reduce individual data variations and enhance bio-signal interpretability.


To model stress levels, we opted for a Multilayer Perceptron (MLP). This MLP is structured with a three-layer design, including hidden layers that consist of 128 and 32 neurons, respectively. It processes an input comprising the 12 extracted features related to heart rate (HR) and heart rate variability (HRV). The output of the MLP is a binary classification, differentiating between states of stress and no stress.



\section{Evaluation}

This section is dedicated to the evaluation of our proposed framework.
We establish a baseline using the pre-trained model to highlight the benefits of incorporating transfer learning. 
Additionally, we explore the delicate balance between privacy preservation and model accuracy by adjusting the privacy budget parameter, $\epsilon$. 

\subsubsection{Implementation Details}


We train the proposed stress detection model by the Adam optimizer (learning rate=$0.001$) and cross-entropy loss function.
A dropout strategy (rate=$0.5$) is implemented after the initial hidden layer to reduce overfitting.
Differential privacy via Laplacian noise addition to the gradients was introduced, with an $\epsilon$ value of $1$, aiming for a balance between privacy protection and model accuracy.

We pre-train on $D_{pre-train}$ with 3271 labeled samples for 50 epochs to capture a wide array of stress signals. Subsequently, we applied FedAvg to fine-tune the model on $D_{fine-tune}$, consisting of 1220 labeled samples, over 30 epochs for personalized user data adaptation. 

The dataset $D_{fine-tune}$ was divided in a manner that allocated the last 30\% of each user's chronologically ordered data to the test set, ensuring the model was trained on past data and evaluated on future data. 
Client partitioning for federated learning was conducted based on individual users, reflecting a real-world scenario where fine-tuning occurs on each user's client device in a personalized manner.



\subsubsection{Results}
The results of our proposed framework for stress detection are shown in Table~\ref{tab:performance_comparison}.
The Plain model, trained solely on the original dataset $D_{fine-tune}$ without incorporating transfer learning techniques, yielded an accuracy of 0.43, an F1 Score of 0.39, a recall of 0.54, and a precision of 0.31. 
Conversely, training the model on the public, more extensive dataset $D_{pre-train}$ resulted in improved metrics: accuracy rose to 0.51, F1 Score to 0.44, recall to 0.25, and precision to 0.36. 
Notably, the integration of transfer learning within our differential privacy federated learning framework enhances performance, achieving an accuracy of 0.53, an F1 Score of 0.52, a recall of 0.75, and a precision of 0.40.


\begin{table}[ht]

\centering
\begin{threeparttable}
\caption{Stress Detection Performance of Different Models}
\label{tab:performance_comparison}
\begin{tabular}{ccccc}
\toprule
\textbf{Model} & \textbf{Accuracy} & \textbf{F1 Score} & \textbf{Recall} & \textbf{Precision} \\
\midrule
Plain    & 0.43               & 0.39               & 0.54                & 0.31              \\
Pre-trained & 0.51       & 0.44                & 0.58                  & 0.36            \\
\textbf{Fine-tuned} & \textbf{0.53}            & \textbf{0.52 }             & \textbf{0.75}              & \textbf{ 0.40}            \\
\bottomrule

\vspace{-10pt}
\end{tabular}



\end{threeparttable}
\vspace{-2mm}
\end{table}

\begin{figure}[t]
\centering
\includegraphics[width=0.42\textwidth]{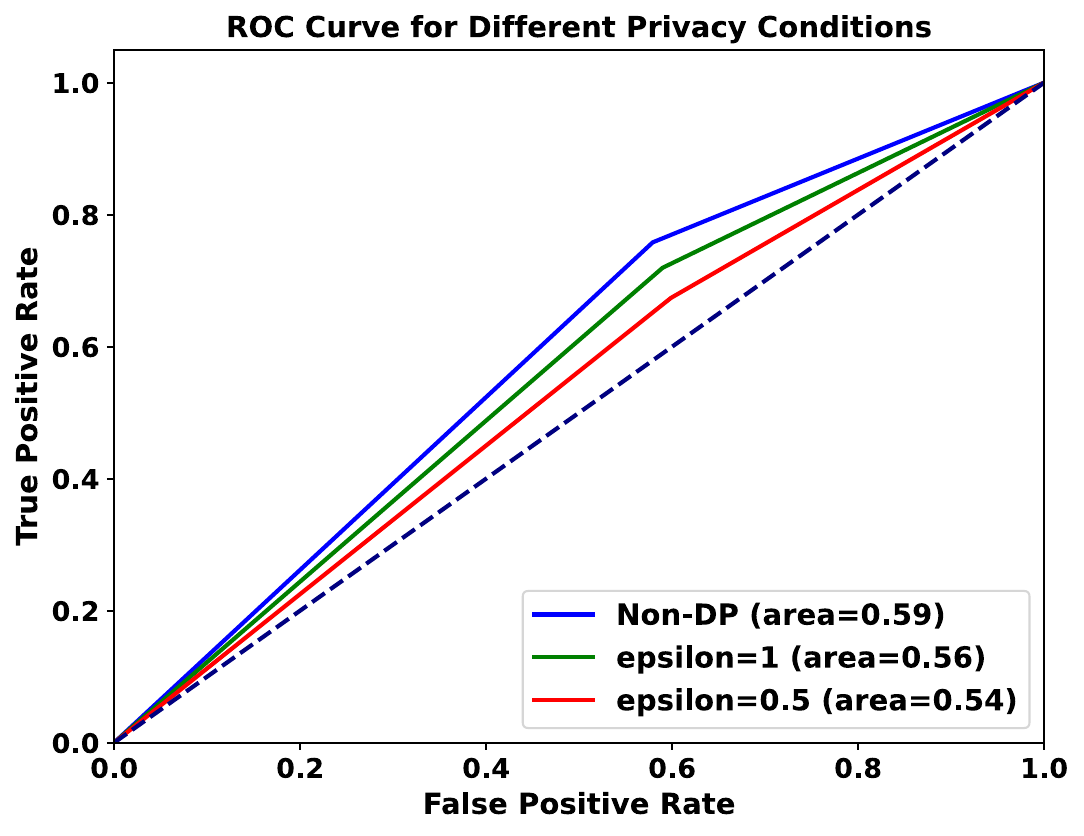}
\vspace{-10pt}
\caption{Overview of the proposed approach.}
\vspace{-15pt}
\label{fig:privacy_roc}

\end{figure}

\subsubsection{Privacy Budget Analysis}

In the final stage of our analysis, we conducted comparisons of our outcomes with two reference benchmarks: one involving the identical neural network framework without the implementation of federated learning or differential privacy, and the other utilizing the same neural network structure with federated learning but absent differential privacy. This comparative evaluation was performed across two scenarios lacking differential privacy, with $\epsilon$ values set at 0.5 and 1, respectively. It's important to note that a lower $\epsilon$ value signifies enhanced privacy safeguards. As depicted in Fig. \ref{fig:privacy_roc}, integrating differential privacy within a federated learning context did not significantly alter the metrics of the receiver operating curve (ROC). This observation suggests that it is feasible to strike an effective equilibrium between preserving model efficacy and upholding privacy standards.

\section{Discussion}
In this study, we introduce a differential privacy federated transfer learning framework and evaluate it on stress detection. 
By applying transfer learning, our approach effectively leverages the strengths of both public and private data sources, enhancing the model's performance in real-world scenarios.
Especially noteworthy is the improvement in recall indicates that our framework is effective in reducing potential overlooking true stress cases.
 This advance is crucial, minimizing missed detections of stress and thereby mitigating their potential adverse effects on individual well-being and public health.
 Additionally, the implementation of differential privacy within the framework ensures the protection of sensitive health data, which is a critical concern in the field of mental health. 
 By infusing noise into the model updates, our framework mitigates risks associated with several cyber attacks~\cite{fredrikson2015model,nasr2018comprehensive,bagdasaryan2020backdoor}.

This study's limitation is its focus on stress detection, which may not encompass other mental health conditions or capture long-term trends effectively due to its design for short-term analysis. Future efforts could aim to broaden the framework's applicability to various mental health states and for different cohorts. This could improve its predictive power for both immediate and extended periods that were shown feasible in prior research~\cite{yang2023loneliness}.

\section{Conclusion}


This study proposed a differential private federated transfer learning framework addressing the challenges of data scarcity and privacy protection in mental health monitoring. 
The proposed framework exhibited strong performance in stress detection while effectively preserving data privacy.
Specifically, our framework achieved a substantial 10\% improvement in accuracy and a noteworthy 21\% enhancement in recall. These results highlighted the potential of our approach to provide more effective and privacy-conscious mental health monitoring, addressing the pressing need for efficient solutions in this domain.

\bibliographystyle{ieeetr}
\bibliography{main}
\end{document}